\documentclass[conference]{IEEEtran}
\IEEEoverridecommandlockouts
\usepackage{graphicx}
\usepackage{siunitx}
\usepackage{booktabs}
\usepackage{tabularx}
\usepackage{multirow}
\usepackage{pifont}
\usepackage[table, dvipsnames]{xcolor}
\usepackage{bm}
\usepackage{bbm}
\usepackage{mathtools}
\usepackage{amsmath}
\usepackage{url}
\usepackage{svg}
\usepackage{times}
\usepackage{amssymb}
\usepackage{url,hyperref}
\usepackage[most]{tcolorbox}
\usepackage[ruled,vlined]{algorithm2e}
\usepackage[framemethod=tikz]{mdframed}
\usepackage{subfigure}
\usepackage{graphicx}
\usepackage{array, boldline, makecell}
\usepackage{todonotes}

\definecolor{lgray}{gray}{0.95}
\newcommand{\commentout}[1]{}
\definecolor{cadmiumgreen}{rgb}{0.0, 0.42, 0.24}
\definecolor{darkgreen}{rgb}{0.3,0.5,0.3}
\definecolor{darkblue}{rgb}{0.3,0.3,0.5}
\definecolor{darkred}{rgb}{0.5,0.3,0.3}

\def\BibTeX{{\rm B\kern-.05em{\sc i\kern-.025em b}\kern-.08em
    T\kern-.1667em\lower.7ex\hbox{E}\kern-.125emX}}

\def\MasMet{\emph{MCE}\xspace }
\def\NNMC{\emph{NN-MC}\xspace }
\newboolean{showcomments}
\setboolean{showcomments}{false}
\ifthenelse{\boolean{showcomments}}
{ }
\begin{document}

\title{End-to-End Learning from Noisy Crowd to Supervised Machine Learning Models
}
\author{\IEEEauthorblockN{Taraneh Younesian\IEEEauthorrefmark{1}, Chi Hong\IEEEauthorrefmark{1}, Amirmasoud Ghiassi\IEEEauthorrefmark{1}, Robert Birke\IEEEauthorrefmark{2} and Lydia Y. Chen\IEEEauthorrefmark{1}}
\IEEEauthorblockA{
\IEEEauthorrefmark{1}TU Delft, Delft, the Netherlands. Email: \{T.Younesian, C.Hong, S.Ghiassi, Y.Chen-10\}@tudelft.nl\\
\IEEEauthorrefmark{2}ABB Research, Baden-D\"attwil, Switzerland. Email: robert.birke@ch.abb.com\\
}
}

\maketitle

\begin{abstract}
Labeling real-world datasets is time consuming but indispensable for supervised machine learning models.  A common solution is to distribute the labeling task across a large number of non-expert workers via crowd-sourcing. Due to the varying background and experience of crowd workers, the obtained labels are highly prone to errors and even detrimental to the learning models. In this paper, we advocate using hybrid intelligence, i.e., combining deep models and human experts, to design an end-to-end learning framework from noisy crowd-sourced data, especially in an on-line scenario. We first summarize the state-of-the-art solutions that address the challenges of noisy labels from non-expert crowd and learn from multiple annotators. We show how label aggregation can benefit from estimating the annotators' confusion matrices to improve the learning process. Moreover, with the help of an expert labeler as well as classifiers, we cleanse aggregated labels of highly informative samples to enhance the final classification accuracy. We demonstrate the effectiveness of our strategies on several image datasets, i.e. UCI and CIFAR-10, using SVM and deep neural networks.
Our evaluation shows that our on-line label aggregation with confusion matrix estimation reduces the error rate of labels by over 30\%. Furthermore, relabeling only 10\% of the data using the expert's results in over 90\% classification accuracy with SVM.

% Acquiring real-world labeled datasets is a time consuming and expensive task. A popular solution to overcome this issue is crowd-sourcing by collection the labels from large-scale distributed workers. However, the labels obtained from the crowd are highly prone to noise due to the lack of expertise of the cheap workers. In this paper we address the issues arising from noisy crowds with various expertise. We summarize the state-of-the-art of noisy crowds and discuss the challenges of label aggregation and learning from multiple annotators from both off-line and on-line data collection viewpoint. We show how benefiting from label aggregation and annotators' confusion matrix estimation improves the learning process from the crowd. Moreover, benefiting from an expert labeler as well as a classifier, we attempt to improve the quality of the labels by further cleaning the aggregated labels of the highly informative data samples to achieve a highly accurate classification model. To demonstrate the effectiveness of the introduced strategies, we test them on several image datasets, i.e. UCI and CIFAR-10, using SVM and deep neural networks. Our evaluations show that the introduced label aggregation and confusion matrix estimations reduces the error rate of the labels over 30\%. Furthermore, relabeling only 10\% of the data using the expert results in over 5\% improvement in classification accuracy.
\end{abstract}

\begin{IEEEkeywords}
crowd-sourcing, label aggregation, active learning, confusion matrix estimation
\end{IEEEkeywords}

\section{Introduction}
%\td{Cite our own last year CogMI paper and your own axive papers, e.g., Masou's TrustNet, and others}
Many artificial intelligence applications rely on supervised learning and  labeled datasets, such as image classification \cite{Wang:2017:IEEETrans:ActiveImageClassification}, activity recognition \cite{Krawczyk:2017:activeRecognition}, and sentiment analysis \cite{Smailovic:2014:InfSci:activeSentiment}. The dataset size and quality directly affect the performance of learned models~\cite{Zhang17:ICLR:memorization} making labeling a daunting task. 
% However, in most of these applications, labeling the data is very expensive and time consuming.
Crowd-sourcing~\cite{Yuen:2011:passat:crowdsurvey} aims to curtail the labeling effort by submitting the data to a large crowd for labeling. Different from traditional labeling campaigns which assume the presence of (few) expensive experts providing the labels, crowd-sourcing relies on several cheap annotators with highly varying knowledge and level of interest~\cite{Howe:2008:crowd}. While the labels can be easily gathered from the crowd, the quality of crowd-sourced labels is still an outstanding issue.%, where the annotators have different expertise and level of interest in the task.

Label aggregation is an efficient method to distill the noise of crowd data by finding the consensus among all workers. % \lc{I don't get thee following sentence}Simply applying traditional supervised learning methods with one expert to the multi annotator problem can be inefficient~\cite{Teh:2010:AISTATS:bit}.
%Several studies try to address this challenge.
% There have been several studies in attempt to introduce solutions for knowledge aggregation from these sources.
The main algorithms in this area 
can be categorized in three directions: \textit{Majority Voting}, probabilistic models via EM algorithms, and \textit{Annotators' Expertise Estimation}~\cite{Georgescu12:CIKM:crowdError}. In Majority Voting, the label with the highest consensus among the workers, is selected as the aggregated label for the data~\cite{Raykar10:journalOfmachine:learningCrowd}.
Although some studies rely more on accurate workers~\cite{Huang:2017:IJCAI:DiverseLabelers}, they require a (small) set of golden standard data with known ground truth labels.
Most studies treat the problem as an unsupervised learning task. EM based studies maximize the data likelihood to infer the unknown true labels~\cite{Raykar10:journalOfmachine:learningCrowd, Demartini12:WWW:ZenCrowd}. Some works also estimate the expertise of workers either via their confusion matrix~\cite{Tanno:2019:CVPR:MultiAnnotatorConfusion, goldberger2017adaptation, song2019selfie} or reliability parameter~\cite{yang:2018:corr:alexa}, as well as the difficulty of items~\cite{Whitehill09:NIPS:vote}. The common objective among them is to infer the true labels, independently from the subsequent supervised learning.  

While these methods try to estimate the true label in an unsupervised manner, they exclude the information in the data samples themselves, e.g., features and informativeness of data. 
% To increase the efficiency, active learning can be used beside crowd-sourcing to select the important samples to be labeled by the workers~\cite{mozafari:2014:lvbd:scaling}. <- used below
Active learning techniques~\cite{settles2009active} are designed to query extra information from an oracle for the data whose (true) labels are not readily available. Such an oracle is assumed to know the ground truth, but at high costs, e.g. a human expert. Hence, only the most informative/uncertain data is queried within a given query budget. The majority of active learning approaches focuses on off-line scenarios with constant budgets, except~\cite{ghiassi2019robust,vzliobaite2013active,Younesian20:arxiv:QActor} that explore active learning on one by one drifting streaming data, however, their focus is on single label scenarios. 

The efficiency of active learning relies on identifying the most informative instances to be labeled. Several measures have been proposed in the active learning literature %studies to identify these samples, such as the uncertainty of the model in the classification of a sample,
e.g., based on class probability~\cite{schohn2000less}, entropy value~\cite{holub2008entropy} or posterior predictive densities~\cite{ijcai2019haussmann}. Moreover, some methods try to identify the samples that cause the highest expected gain in the learning performance once they are labeled~\cite{Fu:2018:sigkdd:errorReduction}.

While crowd-sourcing studies have leveraged informative sample selection~\cite{yang:2018:corr:alexa, Song:2018:Knowl.BasedSyst.:confidenceCrowd, Zhong:2015:AAAI:unsure}, the labeling quality of crowd workers remains a challenge. Usually none of the crowd workers is an expert in the problem field. Hence, it can be beneficial to leverage active learning with an expert labeler to assist the learning process~\cite{Yan11:ICML:activeCrowd}. %\tr{Check the Following}\rb{seems a a bit repetition with the first phrase of two paragraphs above.} 
Moreover, the connection between label aggregation and training classification models seems to neglected in many crowd-sourcing studies, as they only focus on label information and exclude information lying in the data features, where active learning can play an important role.

The prior art in both crowd sourcing and label aggregation focus on off-line scenarios where all the data is available at once. However, in some applications the data is collect the data over time in a streaming setting ~\cite{Murilo2019:kdd:streaming}. The challenges in such on-line setting are small training data in each time step and concept drift~\cite{Lu2020:CoRR:conceptDrift}. The small sample set in on-line scenarios prevents the learning process from convergence especially in deep learning \cite{SahooPLH18IJCAI:fly}. Moreover, concept drift, i.e. the change in the statistical properties of the data, require on-line models to be adaptive to the change \cite{Lu2020:CoRR:conceptDrift}. There are on incremental learning algorithms to train models progressively from new data~\cite{neucom18losing,SahooPLH18IJCAI:fly}. Only few consider noisy stream data~\cite{zhu2006effective,chu2004adaptive}.
However these studies consider ensembles of several classifiers to detect noisy labels which is not scalable to large datasets used in deep learning. %\tr{mention label aggregation in online}

In this paper we bring our end-to-end vision to marry crowd-sourcing with active learning for increased efficiency, e.g. higher accuracy at lower number of queries. \cite{mozafari:2014:lvbd:scaling} has been a pioneer for off-line scenario.
We go beyond by discussing the challenges arising in on-line scenario, where data collection happens continuously, and proposing a solution to  address label aggregation in an on-line manner. We show the gain of human experts in further improving the quality of learning systems
and elaborate the benefit of employing a small clean set of data to estimate the annotators' confusion matrix. Finally, we perform a comparative evaluation against the off-line version of the proposed label aggregation method.

% In this paper we bring a visionary overview of the past and future of studies on noisy crowds guided by human experts. We address the previous works, particularly focusing on label aggregation, active learning with noisy annotators, and various studies on confusion matrix estimation. Furthermore, we go beyond the state-of-the-art and discuss the challenges arising in on-line label aggregation and multiple annotators with various expertise. Ultimately, we address label aggregation in an on-line manner, while data arrives in on-line batches of data; we show the effect of human experts in further improving the quality of learning systems with the on-line aggregated label. Moreover, we elaborate the benefit of employing small clean set of data in the annotators' confusion matrix estimation and compare it with the off-line version of the proposed label aggregation method.

% In this work, we employ active learning to further clean the informative crowd labeled data, to guarantee a high label quality by benefiting the knowledge of an expert. 

%\tr{One paragraph on multi-label problem }

\section{State of the art}
In this section we discuss the state of the art in the area of noisy crowds, dividing them into three categories. First we give an overview of the related works on annotators' confusion matrix estimation, then we review the existing research in offline label aggregation, and in the end, we discuss the works tackling noisy annotators with active learning.
\subsection{Off-line Confusion Matrix Estimation}% - Masoud }
%\td{stress why deep classifier approaches may not render well. The following needs to re-written from a much higher level}
The label confusion matrix is a good indicator to determine the noise pattern and ratio. The diagonal elements represent the probability of the correct label while the
off-diagonal elements indicate the probabilities to flip the correct label with a wrong one. Estimating the confusion matrix can help to correct noisy labels.
Some estimation methods rely on a (small) set of clean samples with known ground truth. For instance, GLC~\cite{hendrycks2018using} estimates confusion matrix assuming a small proportion of trusted data is available.
% Among the various approaches to estimate the confusion matrix, some methods~\cite{hendrycks2018using}, access to a part of clean samples is assumed.
This clean fraction of the dataset improves the estimation accuracy significantly~\cite{Ghiassi20:arxiv:TrustNet,ghiassi2020expertnet}. Furthermore, the study in~\cite{sukhbaatar2014training} approximate the matrix of noisy labels stochastically by using correct labels. In addition, they improve the robustness of DNNs using forward loss correction. %\tr{I put TrustNet here, check}. Other approaches rely on adding extra components such as extra network layers~\cite{patrini2017making}.
% for the approximation of noise patterns.
A few methods approximate the confusion matrix by using Generative Adversarial Networks (GAN)~\cite{thekumparampil2018robustness}. These works try to produce noise similar to the noise pattern in the dataset. The generated data is then used to identify the pattern and estimate the confusion matrix. 

%\rb{Should be summarized better via better grouping and merged with above}\lc{I did not change anything hree}
Some works estimate the confusion matrix by leveraging the prior knowledge in the field. For instance,
Forward~\cite{patrini2017making} assumes a known noise transition matrix/estimates and tries to minimize the distance between classification outputs and transition matrix. Masking~\cite{Han:2018:NeurIPS:Masking} uses human cognition to estimate noise and build a noise transition matrix.  Goldberger et al.~\cite{goldberger2017adaptation} on the contrary, estimate confusion matrix using an additional softmax layer in the DNNs. SELFIE~\cite{song2019selfie} proposes a correction method regarding making high precision for unclean samples, then improves the estimated confusion matrix. The work in~\cite{Tanno:2019:CVPR:MultiAnnotatorConfusion} estimates the annotators confusion matrix and the true labels simultaneously. As the network predicts the true label distribution, one can achieve the estimated noisy labels. Here, we investigate estimating the confusion matrix for multiple annotators, which is not studied well in the prior art~\cite{Tanno:2019:CVPR:MultiAnnotatorConfusion}. %The goal is to find the confusion matrix and the true label distribution in a way that their multiplication results in the observed noisy labels, which can be done by minimizing the cross entropy function between the noisy labels and the multiplication. 
In the case of multi annotators, which there is no access to correct them~\cite{goldberger2017adaptation, song2019selfie}, we can estimate each labeler's confusion matrix. After estimation, and combining each matrix knowledge, we can assess the quality of annotators and correct the noisy labels during training by using them in the loss function optimization.

\subsection{Off-line Label Aggregating}

Different works address label aggregation to distill the true label from redundant noisy labels posed as an unsupervised learning task. They differ in the estimation techniques as well as the latent variables on which the model relies.

One of the earliest works was proposed by Dawid and Skene~\cite{dawid1979maximum}. They use the concept of confusion matrix to model the expertise of labelers estimated via an EM algorithm maximizing the data likelihood. 
% showed an EM algorithm for their label aggregation model.In the algorithm, they made use of the concept of confusion matrix to construct their model. In each iteration of the EM algorithm, all noisy labels should be counted to update the confusion matrix.
BCC~\cite{kim2012bayesian} is a probabilistic graphical model version of Dawid\&Skene's EM. To learn the model parameters, the authors design a Gibbs sampler. During the learning process, the conditional distributions of the model parameters must be computed. This requires traversing all noisy labels in each iteration.
Zhou et al.~\cite{zhou2012learning,zhou2014aggregating} propose a minimax entropy estimator and its extensions to label aggregation. The authors set a separate probabilistic distribution for each worker-sample pair. Zhou and He~\cite{zhou2016crowdsourcing} design a label aggregation approach based on tensor augmentation and completion. In these works~\cite{zhou2012learning,zhou2014aggregating,zhou2016crowdsourcing}, noisy labels are reorganized as a three-way label tensor. %The noisy labels must be aggregated together.
The aforementioned models can be regarded as off-line label aggregation. They target the data at hand and can not readily be adapted to learn incrementally to on-line scenarios.
% They always aggregate the noisy labels of all samples together. In online learning scenarios, we periodically receive noisy labels of new samples. 
Here the data are collected periodically batch by batch. Off-line label aggregation needs to aggregate all batches together to achieve good performance on all labels.
This is time consuming and not scalable, because the off-line model must wait to have all data at once to start or retrain on the whole accumulated data at each new batch arrival. %along with the old received batches.
Researchers~\cite{jongen2014relationship,pencavel2015productivity} have demonstrated that people's attention, fatigue and behaviors change over time. Therefore, we need on-line label aggregation algorithms which can continuously update the aggregation model according to the new observed labels  to accurately infer the true labels.

\subsection{Active Learning from Multiple and Noisy Annotators}

Active learning aims to identify informative and representative unlabeled data samples and label them by an expert to increase the efficiency of the training procedure~\cite{settles2009active}. Traditional active learning methods consider an oracle knowing the ground truth for all the data readily available during the learning process~\cite{Li:2013:cvpr:adaptiveActive}. However, this assumption does not apply to real-world applications. A common solution is employing several labelers, weak or strong, in the form of crowd-sourcing. Therefore, leveraging active learning in crowd-sourcing has become an interesting topic. % that has been studied in several works \cite{Yan:2011:ICML:activeCrowds}.

While~\cite{Chaudhuri:NIPs16:ImperfectLablers} considers imperfect labelers that may abstain from labeling, \cite{Huang:2017:IJCAI:DiverseLabelers} assumes having multiple labelers with different costs and qualities. It actively selects both samples and labelers considering sample usefulness and labeler's accuracy and cost, assuming that all the labelers are prone to make mistakes. \cite{Sheng:2008:SIGKDD:Multilabeler} focuses on the selection of informative samples in the presence of several non-expert labelers via majority voting of the labelers. Considering the same framework, an extension to unbalanced labels is studied in~\cite{Zhang:2015:TransCybernetics:UnbalancedMultiLabeler}. However, they fail to leverage the labelers based on their expertise in labeling.
In contrast,
\cite{Tanno:2019:CVPR:MultiAnnotatorConfusion} considers several noisy annotators with unknown expertise and jointly estimates the confusion matrix of the annotators and the true label distribution by minimizing the cross entropy function between predicted noisy labels and given noisy labels. To estimate the workers' expertise, \cite{yang:2018:corr:alexa} uses a low rank representation for the workers' skills and estimates this representation using EM algorithm. A bayesian neural network is used in this study to model the uncertainty in the data to choose the most uncertain samples for labeling by the expert crowd..
Asking the workers to provide their confidence level while labeling has been studied in~\cite{Song:2018:Knowl.BasedSyst.:confidenceCrowd}, although relying on the user provided information seems challenging. Similarly, \cite{Zhong:2015:AAAI:unsure} asks workers to chose the option \textit{unsure} if applicable, which is similar to the abstention of labelers in \cite{Chaudhuri:NIPs16:ImperfectLablers}. Moreover, a few studies consider annotators with various costs and adjust their active learning algorithm to select annotators with balanced cost/accuracy~\cite{Zheng:2010:ICDM:ActiveCost, Donmez:2008:CIKM:proactiveCost, younesian2020active}.

\section{Challenges}
In this section, we discuss the challenges arising from label aggregation for online data streams with multiple annotators.

\subsection{On-line Label Aggregating}% - Hong}

On-line learning requires label aggregation algorithms to continuously aggregate the labels of a data stream. %We assume that the samples and their noisy labels by the crowds are coming batch by batch. 
Because of storage limits, regulation constraints or other factors, data batches are available for a limited duration. Therefore on-line aggregating is different from traditional label aggregation tasks which process all observed noisy labels at once. Therefore, the design of a new learning framework for on-line scenarios is essential. 

The first challenge is how to make use of the knowledge of the old received batches when aggregating a new observed batch. The knowledge of the old batches is valuable for the label aggregation algorithm to precisely evaluate the behaviors of the non-professional crowd workers. However, the old batches are missing when we get the new batch. So our model must be capable to continuously update its parameters according to the knowledge learned from every observed batch.

The second challenge is to design a suitable learning method and optimization goal for the aforementioned application scenario. EM algorithms~\cite{dawid1979maximum}, Gibbs samplers~\cite{kim2012bayesian} and tensor completion methods~\cite{zhou2016crowdsourcing} have been proposed for label aggregation. However, they usually need to travel and count all noisy labels in each iteration. Therefore, these methods are not applicable for the on-line data arrival setting. The challenge then is to find an optimization method which can update the label aggregation model in the presence of a data stream. Besides, it is vital to design a reasonable optimization goal for the model to accurately aggregate the observed noisy labels.

\subsection{Multiple Annotators}% - Taraneh}
Having multiple annotators with different expertise rises the question of which annotator to choose for labeling each data sample. Although methods like label aggregation try to overcome this issue by combining the opinion of all annotators, there is still no guarantee that the aggregated labels are accurate. The challenges are: annotators having different level of knowledge for the task, some annotators could be malicious or simply not willing to put effort for the task, or there could be a relation between the data category and the annotator's expertise level \cite{Teh:2010:AISTATS:bit}. Also, there might be some prior knowledge about each annotator, their skill and cost \cite{Zhong:2015:AAAI:unsure,Song:2018:Knowl.BasedSyst.:confidenceCrowd, Chaudhuri:NeurIPs15:StrongWeakLablers5}. As mentioned earlier, majority voting is one of the simplest ways to combine the annotators' knowledge. However, in difficult cases where most of the annotators could make mistakes, majority voting or other label aggregation methods can fail~\cite{Zhao:PASSAT11:IncRelabel}. Another category of methods tries to estimate the expertise of the annotators to select the most skilled ones~\cite{Teh:2010:AISTATS:bit, yang:2018:corr:alexa}. Taking steps further, selective majority voting~\cite{Yang:SIGKDD12:schoolOFthoughts} applies majority voting to the $D$ most reliable voters based on their estimated expertise. These views, fail to consider the difficulty of the samples as well as the change in the expertise over time. 

One could use an expert opinion to verify the aggregated labels. However, since the expert opinion is expensive~\cite{Chaudhuri:NeurIPs15:StrongWeakLablers5}, another challenge is to reduce this cost by efficiently choosing important samples to be relabeled by the expert~\cite{Chaudhuri:NeurIPs15:StrongWeakLablers5}. Furthermore, as mentioned above, in the cases where there is a relation between the annotators' expertise and the data sample, it is vital to identify those samples for further investigation. In this case, expert knowledge could be used to label these informative samples~\cite{Liu:2015:SIGMETRICS:activeImproving}. 

% \begin{itemize}
%     \item difficult to estimate the expertise
%     \item no guarantee that they will be accurate all the time
%     \item efficiency is sample selection 
% \end{itemize}

\section{Deep off-line label aggregation}
In this section we introduce a method to estimate the confusion matrix of the workers, to select high quality labeler in an off-line manner, benefiting from a small clean set with known true labels.

\subsection{Off-line Multi-Annotator Confusion Matrix Estimation: \emph{\MasMet}}%- Masoud}

In many real-world applications, a small set of correctly labeled data is available. In these cases, an effective estimation method is to extract the annotators' confusion matrix probabilities by using a small proportion of trusted data.

Consider that each image in dataset $\boldsymbol{x}_i \in \mathbb{R}^{n \times m}$ has a set of of labels from different annotator $ \mathcal{Y}_i = \{y_i^1, y_i^2, \dots, y_i^K\}$ where $y_i^{(k)} \in \{1, 2, \dots, C\}$ denotes the $i^{th}$ annotated label from $k^{th}$ annotator. Also, $n$, $m$ and $K$ are number of images, features and annotators, respectively. The confusion matrix of each annotator $C^{(k)}$ is estimated by training a DNN on the dataset $\mathcal{D} = \{(\boldsymbol{x}_1, \mathcal{Y}_1), \dots, (\boldsymbol{x}_n, \mathcal{Y}_n)\} $ which is labeled by annotators. 

\subsubsection{Confusion Matrix Estimation}
To achieve a robust DNN training with images labeled by multiple annotators, we leverage additional information from a small set of clean samples to estimate the confusion matrix which is introduced by~\cite{hendrycks2018using}. The noise confusion matrix guides DNNs to recover the true label of each image. They can either derive the true labels directly, then train DNNs with new cleansed data or correct the loss function implicitly. 
The proposed method by~\cite{hendrycks2018using} starts with training an image classifier on the noisy label data. We train $f_{(k)}(\cdot, \Theta)$ on the dataset $\mathcal{D}_{(k)} \subset \mathcal{D}$ which denotes the dataset annoatetd by annotator $k$. In other word, each annotator generates $\mathcal{D}_{(k)} = \{(\boldsymbol{x}_1, {y}^{(k)}_1), \dots, (\boldsymbol{x}_n, {y}^{(k)}_n)\} $ for training corresponding DNNs $f_{(r)}(\cdot, \Theta)$. After training each network, the elements of confusion matrix $C^{(k)}_{i,j}$ are approximated via a small fraction of trusted data ${\mathcal{D'}}$ including true label ${y'}$. Given $A_i \subset {\mathcal{D'}}$ the subset of trusted data, each elements of $C^{(k)}_{i,j}$ can be estimated by:
\begin{equation}\label{C_est}
    \hat{{C}}^{(k)}_{i,j} = P(y^{(k)} = j | y' = i) \approx \frac{1}{|A_i|} \sum_{x \in A_i} f_{(k)}(y^{(k)} = j | \boldsymbol{x}, \Theta)
\end{equation}
where $f(y^{(k)} = j | \boldsymbol{x}, \Theta)$ denotes the probability of predicted label of $\boldsymbol{x}$ having class $j$. Hence, estimated confusion matrix $\hat{{C}}^{(k)}_{i,j}$ is the mean predicted probability of class $j$ for true label of class $i$ for the trusted data samples. The estimation depends on the annotator's skills and the number of clean labels in trusted data for each class. Annotators skills include the number of incorrect labels assigned to each instance, also in many cases, the pattern of wrong labels follows a specific transition function. 

As mentioned in Eq.~\ref{C_est}, a trained DNN is used to extract the elements of the confusion matrix. In this method, not only the quality of input dataset is essential, but also the architecture of DNN plays a crucial role in approximating a useful noise confusion matrix. %To evaluate our method, we consider an 8-layer CNN with 6 convolutional layers followed by 2 fully connected layers with ReLU activation functions as in~\cite{wang2018iterative}. Even for more accurate estimation, ResNet and Wide-ResNet~\cite{Zgoruyko:BMVC16:WideRestNet} are the two most accurate image classifiers, but their training time is extremely long. Since the main idea of this approach is to train a DNN for each annotator, we decide to employ a CNN for saving training time and meeting the required accuracy. 
 
\subsubsection{Multiple Annotators Multiple Confusion Matrices (\emph{\MasMet})}
After estimating the noise confusion matrix, we need to find the best annotator among them. $\hat{C}^{(k)}$ is our metric to identify the most accurate one for each annotator $k$. The diagonal elements of the matrix $\hat{C}^{(k)}_{i,i}$ indicate the probability that a label is correctly annotated. Generally, we can find the least noisy datasets based on the confusion matrix by calculating the average of \textbf{\textit{trace(.)}} for each matrix $\hat{C}^{(k)}$. We first define a set $\mathcal{T}$ consisting of the average value of \textbf{\textit{trace(.)}} for each $k$. The set can be written as:
\begin{equation}
    \mathcal{T} = \{\frac{1}{C}{\boldsymbol{trace}(\hat{C}^{(1)})}, \frac{1}{C}\boldsymbol{trace}(\hat{C}^{(2)}), \dots, \frac{1}{C}\boldsymbol{trace}(\hat{C}^{(K)})\}
\end{equation}
where $\boldsymbol{trace}(\hat{C}^{(k)}) = \sum_{i=1}^{C}{\hat{C}^{(k)}_{i,i}}$. To choose the most clean dataset, we consider $\mathcal{T}$ as a reference to illustrate average noise ratio for each annotator. The selected dataset can be written as the following:
\begin{equation}
    \mathcal{D}_S = \{(\boldsymbol{x}_i, y^{(k)}_i) | k = \boldsymbol{index}(\min_{j \in \mathcal{T}}{j})\} 
\end{equation}
where $\boldsymbol{index}(.)$ describes the index of an element in a set. Next, we can train a DNN with the selected dataset $\mathcal{D}_S$, which contains less corruption than other labeled datasets. In other word, we choose the most accurate annotator, and as a result the obtained labels will be more reliable than the rest.
The aforementioned method works based on the diagonal elements in each annotator's confusion matrix.

\section{Proposed on-line label aggregation}
%\rb{Introduce what \NNMC is and what it stands for and its acronym}
%\tr{Check Following}
%Although benefiting from the small clean labeled set is effective to accurately estimate the confusion matrices, this labeled set might not be available in all applications. 
With increasing practise of on-line data curation, the label set is continuously updated or may not be possible to store for all applications.
Moreover, training a neural network can be expensive in some cases and requires the dataset to be available ahead of time. Intelligent selection of the clean labeled data could be more efficient for the learning process.
In this section, we introduce a novel end-to-end framework to aggregate labels of the data annotated by crowd workers in an on-line streaming data setting. To further improve the quality of the aggregated labels, benefiting from a classifier, we leverage active learning to cleanse/relabel informative samples by an expert and train the model on high quality data. %Moreover, benefiting a set of ground truth labels, we introduce a method to estimate the confusion matrix of the annotators and compare it with our online label aggregation methodology.   
\subsection{Problem Definition}

We focus on an on-line data arrival setting that consists of two steps: \textit{i}) label aggregation, and \textit{ii}) active learning. %On the contrary, the second scenario is off-line, where the whole dataset is available ahead of time.  
Consider data which periodically streams into the classifier in small batches $D$ for training. The instances of the training data are labeled by the crowd, as each instance takes the form $(\bm{x}_j, \tilde{y}_{j,1}, ..., \tilde{y}_{j,K})$, where $\tilde{y}_{j,k}$ means the potentially noisy label provided by worker $k$ for sample $j$. $\bm{x}_j$ represents the feature inputs. Therefore, we have the feature inputs together with multiple potentially noisy labels in an instance. Our task is to train a classifier with this data stream.

\begin{figure*}[t]
\begin{center}
   \includegraphics[width=0.7\textwidth]{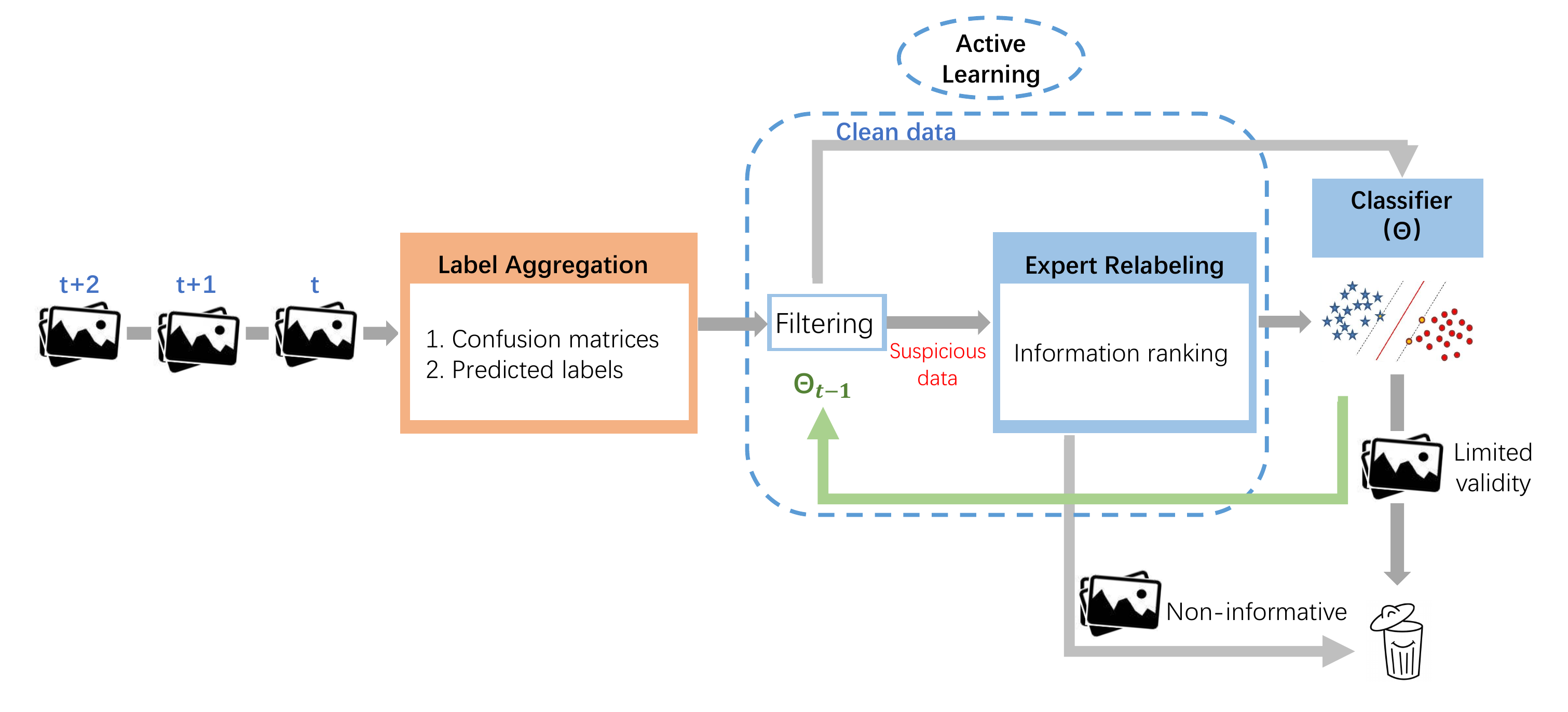}
\end{center}
\caption{The Workflow of the End-to-End On-line \NNMC with Active Learning}
\label{workflow}
\end{figure*}

The data stream will be processed by label aggregation first. The label aggregation algorithm can give an aggregated label (the predicted true label) $\tilde{y}_j$ for each instance $j$ according to the corresponding noisy labels $\bm{Y}_j = \{\tilde{y}_{j,1}, ..., \tilde{y}_{j,K}\}$. The label aggregation aims to lower the noise rate in the labels. However, since the label aggregation algorithm is not perfect, the aggregated labels can be wrong. Therefore, active learning is essential to clean the aggregated labels. In contrast to label aggregation, in active learning, the information of the machine learning classifier is also used to detect informative/useful samples to relabel. Note that it is expensive to verify every sample because of the limited budget. Therefore, the label aggregating process before active learning is useful to increase the quality of the labels. The goal of the active learning step is to identify samples potentially mislabeled by the label aggregation step and relabel them by an oracle to reach the ultimate goal of high classification accuracy. Finally, the classifier will be trained by high quality data in a supervised manner. Figure~\ref{workflow}
shows the full framework of our algorithm. The details of each step will be discussed in the following sections.

\subsection{On-line Label Aggregating}% - Hong}

\subsubsection{Basic Setting}

In our online learning setting, %the data periodically stream into the classifier in small batches $D$ for training. The form of a data sample is $\{\bm{x}_j, \tilde{y}_{j,1}, ..., \tilde{y}_{j,K}\}$, where $\tilde{y}_{j,k}$ means the potentially noisy label provided by worker $k$ for sample $j$. %Then,
the main task of label aggregating is to choose one label $\tilde{y}_j$ for each sample $j$ according to its noisy labels $\bm{Y}_j = \{\tilde{y}_{j,1}, ..., \tilde{y}_{j,K}\}$. In this online case, for each batch of samples, our label aggregation model will give the corresponding aggregated labels in real time.

\subsubsection{Algorithm Framework}

We use $p$ to denote our label aggregation model. In order to update the model in our online learning setting, we use stochastic optimization methods like SGD and RMSProp~\cite{tieleman2012lecture}. These methods are easy to apply for mini-batch learning which fits our on-line learning setting well. Then, we need to choose an optimization goal for the optimization method. Our goal is to maximize the data likelihood of the noisy labels. The optimization function of this algorithm is designed according to variational inference. We use an implicit distribution $q$ as the approximate distribution. Then we set minimizing the Kullback-Leibler divergence between $p(\tilde{\bm{y}}|\bm{Y})$ and $q(\tilde{\bm{y}}|\bm{Y})$ as the optimization function, where minimizing the Kullback-Leibler divergence is equivalent to maximizing the evidence lower bound of the log data likelihood $\log p(\bm{Y})$.

\subsubsection{Neural Network based Multi Class Aggregation (NN-MC)}
According to the algorithm framework, we can define our model by specifying the forms of $p$ and $q$. In order to apply stochastic optimization methods in the label aggregation model, the loss function must be differentiable with respect to the model parameters of $p$ and $q$. The definition of our \NNMC model is discussed below. $p$ is defined using the concept of confusion matrix~\cite{dawid1979maximum}. $C^{(k)}$ represents the confusion matrix of worker $k$, where its element $C^{(k)}_{i,j}$ is the probability that worker $k$ gives a label $j$ when the true label of the item is $i$. That is to say, $C^{(k)}_{i,j} = p(\tilde{y}_{j,k} = c | y_j = t)$. In \NNMC, $q$ is a neural network which represents a distribution $q(\tilde{y}_j|\bm{Y}_j)$. Then, according to the definition, we can calculate the loss function and apply mini-batch stochastic learning for \NNMC.

In on-line learning scenarios, at the beginning, \NNMC uses the noisy labels of some samples to initialize the model parameters. After the initialization, the data batches will be input into \NNMC one by one. For each batch of noisy labels, \NNMC uses them to update the model parameters (e.g., confusion matrices, neural network parameters) and then estimates the most confident labels (aggregated labels) for the corresponding samples. After learning the values of the confusion matrices, it is easy to compute the aggregated labels by maximizing the data likelihood of the observed noisy labels. \NNMC can also be applied to off-line cases by updating the model parameters with all noisy labels and aggregating all the noisy labels using the learned parameters together.

It should be note that as the introduced off-line confusion matrix estimation \MasMet aims to estimate the confusion matrices, we can modify the confusion matrix estimation step of the \NNMC based on the proposed \MasMet, and leverage the rest of \NNMC approach by using maximum likelihood to extract the aggregated labels. In other words, in the applications where a small clean data is available, \MasMet can assist \NNMC to estimate the expertise of the workers, however, only in an off-line setting.

\subsection{Active Label Cleansing}% -Taraneh}
After getting the aggregated labels for each batch of data, we aim to use the expert knowledge to further cleanse the potential wrongly aggregated labels, i.e. noisy labels. This step uses the relationship between a classifier and the features of the data samples. Our framework is based on streaming data where the data arrives in small batches, is used in the training process and then discarded. Each data instance $(\mathbf{x}_j, \tilde{y}_j)$ in the upcoming batch $D$ contains feature inputs $\boldsymbol{x}_j \in \mathcal{X} \subset \mathbb{R}^m$ and a potentially noisy aggregated label $\tilde{y} \in \mathcal{Y} := \{1, ..., N\}$. The goal is to relabel informative wrongly annotated samples by their true label $y$. Our algorithm consists of three steps: \textit{i) filtering}, \textit{ii) informative sample selection}, and \textit{iii) relabeling}. %We will explain each step as following:

\subsubsection{Filtering}
The first step is to identify the samples that have been annotated with a wrong label during the label aggregation process. One way is to leverage the classifier's prediction. By comparing the classifier's prediction $\hat{y}_j$ with the aggregated label $\tilde{y}_j$, we consider a sample to be clean if the predicted label and the aggregated label are the same, i.e. $\hat{y}_j=\tilde{y}_j$, and add them to the clean set $C=\{(\mathbf{x}^c_j,y_j)\}$. The rest of the samples are considered suspicious $U=\{(\mathbf{x}^u_j,\tilde{y}_j)\}$. 

\subsubsection{Informative Sample Selection}
The next step is to identify the informative samples among the suspicious set $U$ to query their true label from the expert. The purpose of this step is to avoid the cost of relabeling the whole suspicious set, due to the expensiveness of the expert. We use two methods to measures informativeness and rank the samples: \emph{Least Confident(LC)} and \emph{Best-versus-second-best (BvSB)}~\cite{joshi:2009:CVPR:multi}. Both of these methods consider the samples highly informative, if the classifier's uncertainty in their classification is high. Consider the classifier's prediction probability vector for the data sample $\mathbf{x}_j$ as $\mathbf{p}(\mathbf{x}_j)$, therefore, $p_{best}$ and $p_{second-best}$ represent the most likely and the second most likely class to assign for that data sample. LC compares samples based on how least confident the model is to classify them, i.e. $I(\mathbf{x}_j)= p_{best}(\mathbf{x}_j)$. Whereas, BvSB compares samples based on how much the model is confused between the two most probable classes, i.e. $ I(\mathbf{x}_j)=p_{best}(\mathbf{x}_j)-p_{second-best}(\mathbf{x}_j)$.
The value $I(\mathbf{x}_j)$ shows the informativeness of the data sample $\mathbf{x}_j$.
The lower the $I(\mathbf{x}_j)$ is, the more difficult and confusing the sample is, therefore the sample is highly informative and useful to be relabeled. 

To select highly informative samples, we rank them based on their $I(\mathbf{x}_j)$ value in an increasing order.

\subsubsection{Relabeling}
After ranking the samples based on their informativeness, we select the top $r$ samples to relabel by the expert labeler, i.e. the oracle. We add the relabeled clean samples to the samples filtered as clean in the filtering step and re-train the classifier with the clean dataset for the current batch. This process is repeated at each batch arrival.

%In many real-world applications, collecting a dataset with correct labels is impossible because it is costly and time-consuming. Hence, datasets were obtained from cheap crowdsourcing and annotators. The skill of each annotator is different based on the pattern and ratio of noise. Thus we can estimate the noise confusion matrix for each of them. The noise confusion matrix contains the alteration probability of one label to another label that it helps to be aware of noise characteristics and correct the wrong labels. 

\section{Preliminary Evaluation}
In this section we evaluate our off-line confusion matrix estimation \MasMet, as well as the proposed end-to-end on-line \NNMC with active learning, on two classifiers. First, we compare our proposed off-line confusion matrix estimation and label aggregation methods using convolutional neural networks. Second, we present our experimental results for end-to-end on-line label aggregation with active learning using SVM. 
\subsection{Experimental Setup}
\subsubsection{Datasets}
We evaluate the introduced frameworks on two types of datasets. The first type represents less complicated smaller sized data, with fewer and handcrafted features that are suitable to train standard ML approaches. The second type instead uses directly the pixels values and represents the deep learning approach which integrates feature selection into the training process.
For the first type we use four multi-class datasets with different sizes and features from the \emph{UCI machine learning repository} \cite{Dua:2017}: \emph{letters}, \emph{pendigits}, \emph{usps} and \emph{optdigits}. The \emph{letters} dataset tries to identify the 26 capital letters of the English alphabet with 20 different fonts. The remaining three target the recognition of handwritten digits via different handcrafted features and from different number of people. These datasets are used to evaluate \NNMC and active relabeling. 
For the second type we use the well-known CIFAR-10 dataset~\cite{kriz-cifar10}. This dataset consists of colored $32\times32$-pixel images divided into ten categories. This dataset is used for the comparison between \MasMet and \NNMC for confusion matrix estimation. This dataset is selected since \MasMet uses deep neural networks that are successful in classifying more complex datasets like CIFAR-10.
Table~\ref{tab:datasets} summarises the characteristics of both groups of datasets.
Since these datasets contain only one label per data, we need to synthesize the noisy crowd, where each worker assigns a noisy label to the data using the procedure in following section.

\begin{table}[t]
\centering
\caption{Summary of the main properties of evaluated datasets.}
%\rb{uncommented the train and test rows}}
\label{tab:datasets}
\resizebox{\columnwidth}{!}{
\begin{tabular}{l|cccc|c}
\hline
\rowcolor{gray!25} \textbf{Dataset}      & \textbf{letters} & \textbf{pendigits} & \textbf{usps} & \textbf{optdigits} & \textbf{CIFAR-10} \\ \hline
\# classes  $~k$  & 26  & 10      & 10   & 10  & 10     \\
\# features $d$ & 16 & 16         & 256  & 64    & 32x32x3    \\
\# train    & 15000  & 7494       & 7291 & 3823 & 50000     \\
\# test     & 5000   & 3498       & 2007 & 1797 & 10000  \\ 
\hline
\end{tabular}
}
\end{table}

\subsubsection{Annotation Noise}
To model imperfect annotators (workers), we use four noise pattern with noise rate $\varepsilon$. Worker $k$ with noise rate $\varepsilon$ assigns the ground truth label for each data point with probability of $1-\varepsilon$ and makes a mistake (assigns another class) with  probability $\varepsilon$. The wrong class can be selected in various ways that are associated with different noise patterns as follows:

\begin{itemize}
    \item \textit{truncnorm}: uses a truncated normal distribution $\mathcal{N}^T(\mu, \sigma, a, b)$ motivated by~\cite{goodfellow:2015:explaining}. We scale $\mathcal{N}^T(\mu, \sigma, a, b)$ by the number of classes $C$ and center it around a target class $\tilde{c}$ by setting $\mu = \tilde{c}$ and use $\sigma$ to control how spread out the noise is. $a$ and $b$ simply define the class label boundaries, i.e. $a=0$ and $b=C-1$. We set $\mu=3$ and $\sigma=1$ in our experiments.
    
    \item \textit{bimodal}: is an extension of \textit{truncnorm}. This pattern combines two truncated normal distributions. It has two peaks in $\mu_1$ and $\mu_2$ with two different shapes controlled by $\sigma_1$ and $\sigma_2$. The peaks are centered on two different target classes $\mu_1 = \tilde{c_1}$ and $\mu_2 = \tilde{c_2}$. We use $\mu_1=3$, $\sigma_1=1$, $\mu_2=7$, and $\sigma_2=0.5$.
    
    \item \textit{flip}: considers partial targeted noise where only a subset of classes, $\{2, 3, 4, 5, 9\}$ in our example, are affected by targeted noise, i.e. swapped with a specific other class~\cite{wang2019symmetric}.
    
    \item \textit{uniform}: uniformly selects one of the wrong labels.  
\end{itemize}

\begin{figure*}[t]
\begin{center}
\subfigure[Bimodal]{
    \includegraphics[scale=0.3]{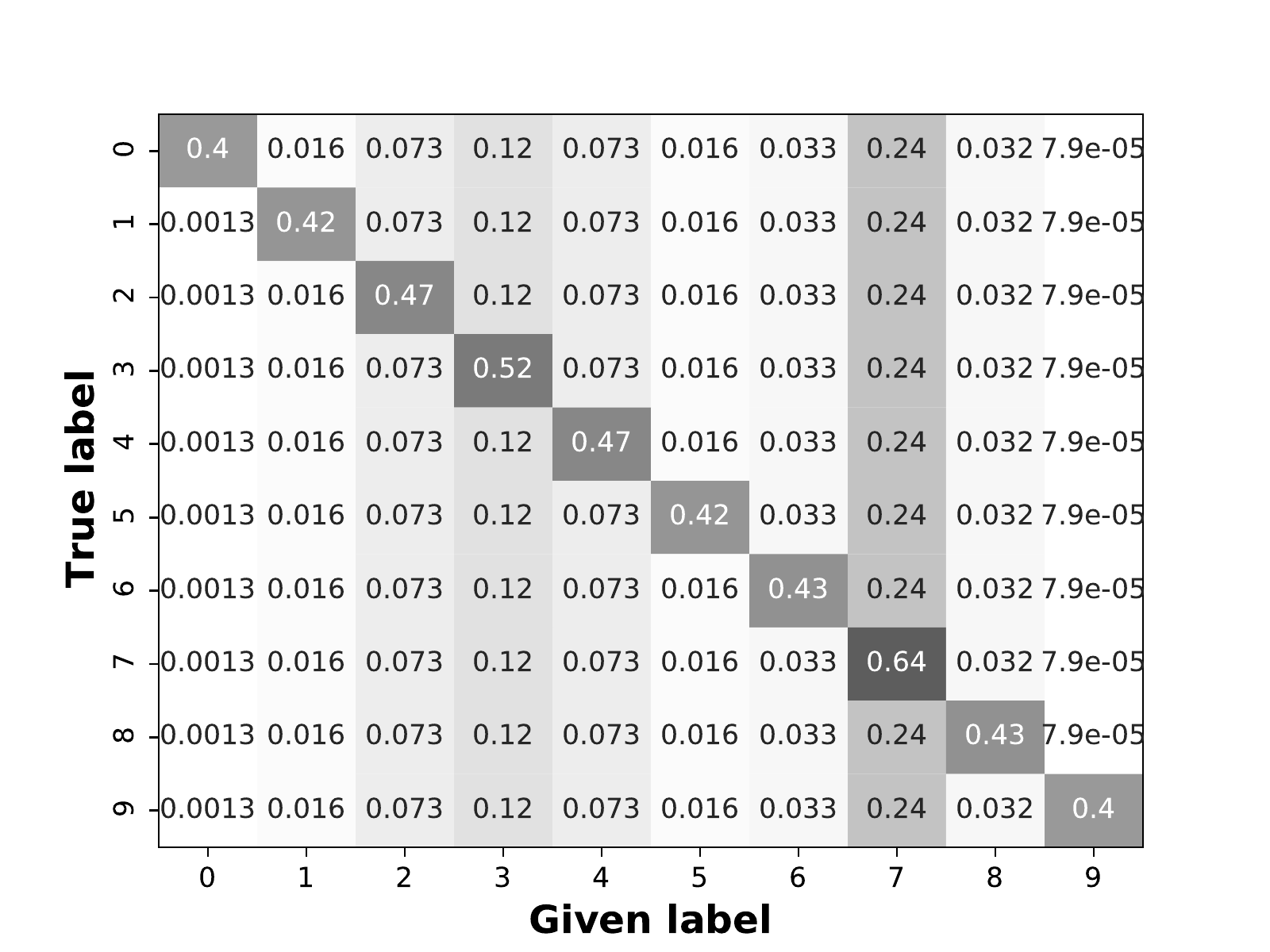}
}
\subfigure[Truncnorm]{
    \includegraphics[scale=0.3]{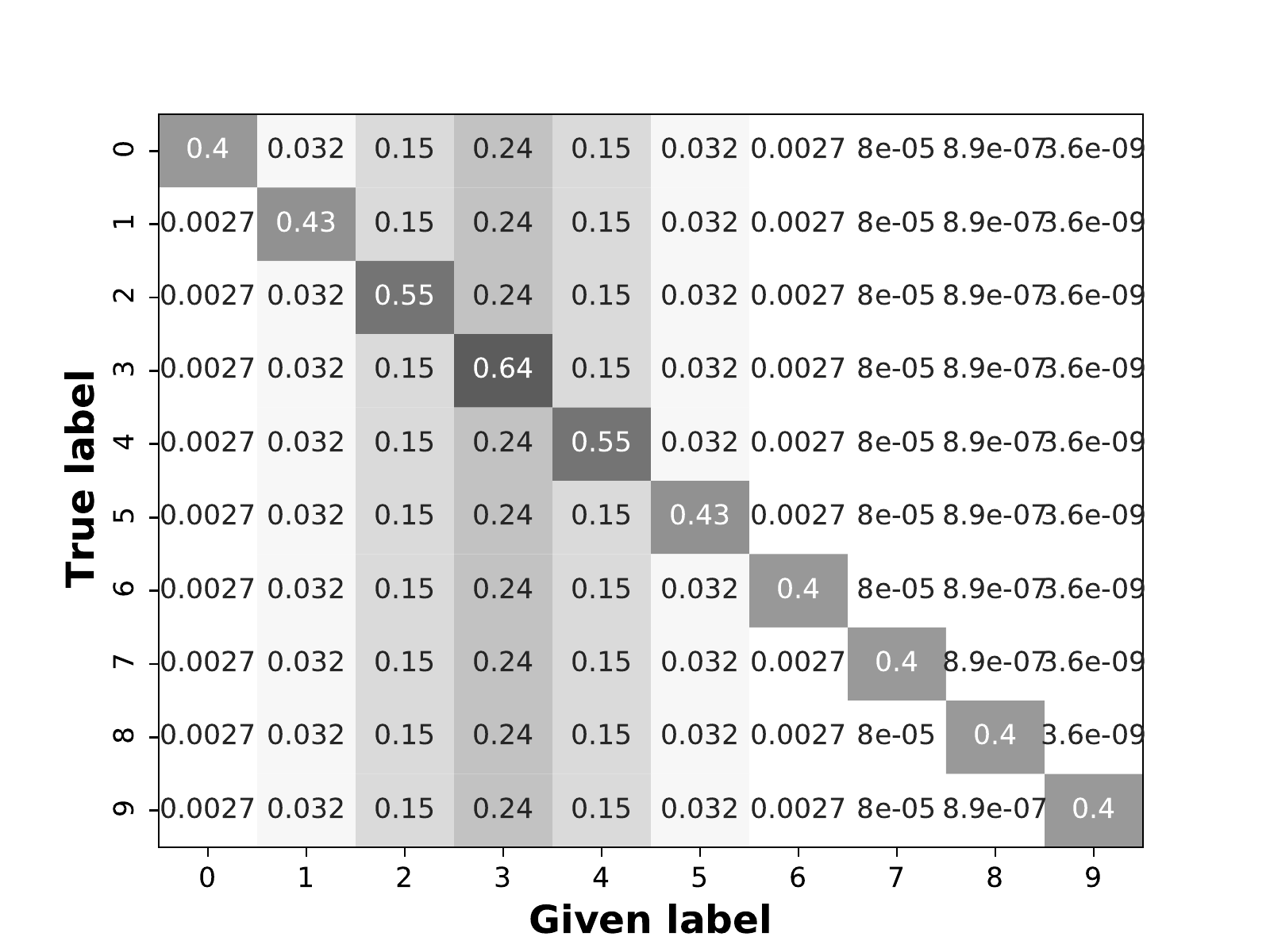}
}
\subfigure[Flip]{
    \includegraphics[scale=0.3]{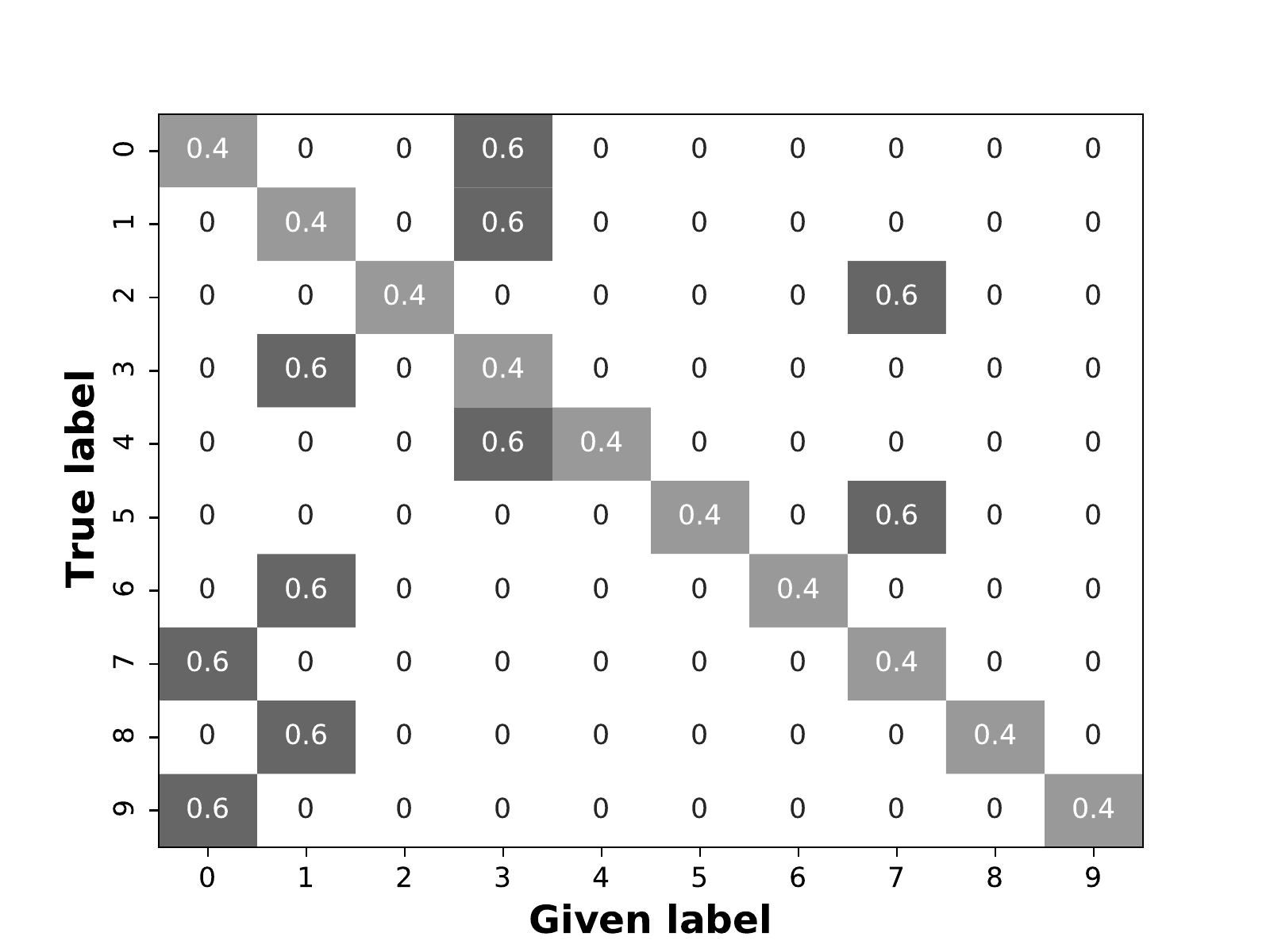}
}
\subfigure[Uniform]{
    \includegraphics[scale=0.3]{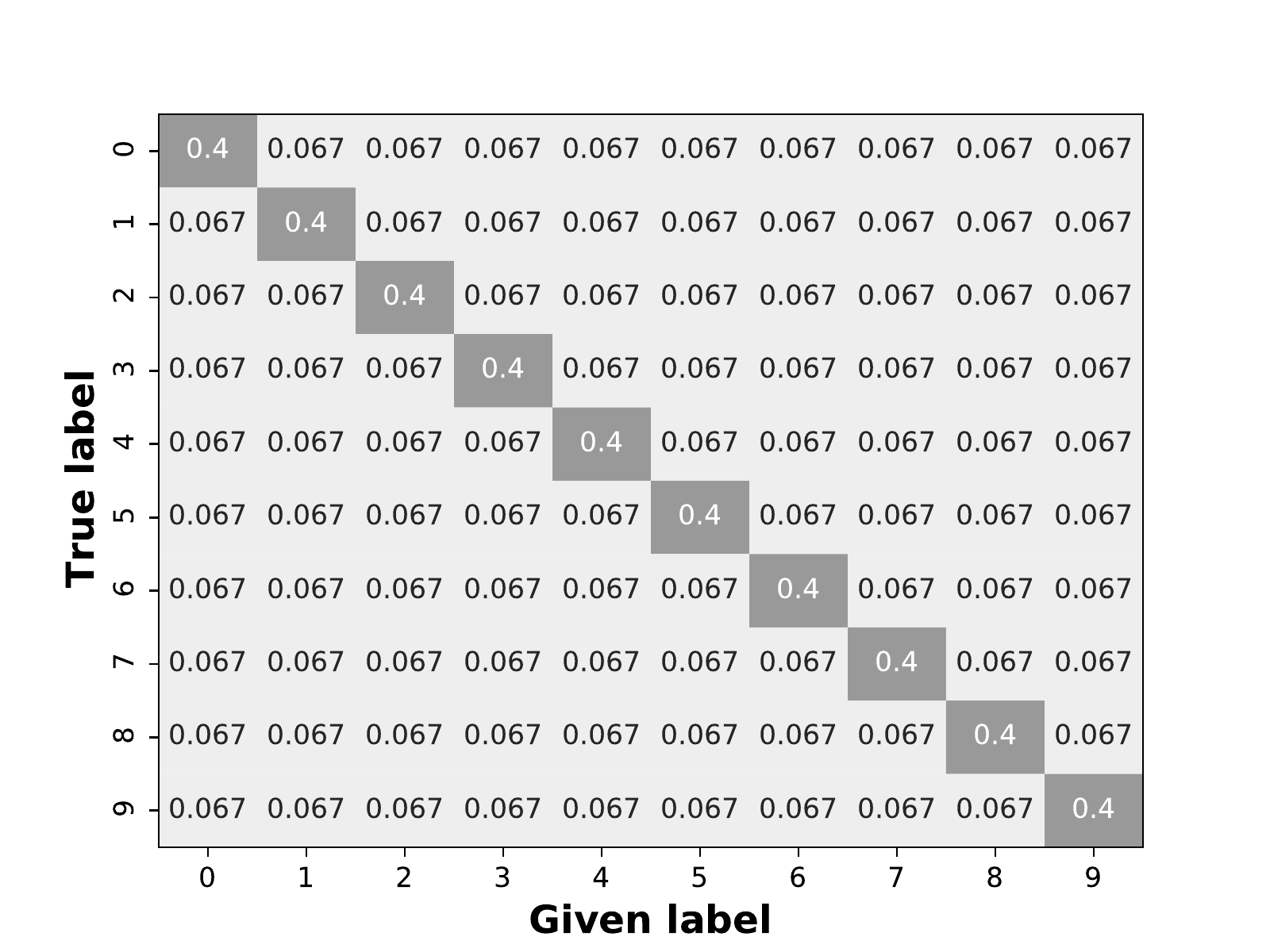}
}
\end{center}
\caption{Generated confusion matrices, where the number of classes is 10 and $\varepsilon=0.6$}
\label{cmatrix}
\end{figure*}

Figure~\ref{cmatrix} illustrates the confusion matrix of an annotator with the introduced noise patterns using $\varepsilon=0.6$.

% \begin{figure*}
% \begin{center}
%   \includegraphics[width=\textwidth]{cifar10-ana2d.pdf}
% \end{center}
% \caption{Error rates for on-line and off-line \NNMC, and Majority Voting for CIFAR-10. At each row, one parameter is changing while other parameter values are fixed. Parameters are fixed as number of workers on $K=6$, empty proportion on $e=0.1$, noise rate on $\varepsilon=0.6$, noise patter on bimodal}
% \label{cifar}
% \end{figure*}

\subsubsection{Training Parameters for UCI Datasets}
As \NNMC needs an initial training phase, we use an initial set of 50 samples for all datasets except \emph{letters}. For \emph{letters}, since the number of classes is higher and the dataset is more complex, we let the initial set be $150$ samples. The instances are chosen randomly from the training set. To speed up training, we limit the datasets to the total training size of $N=1050$ (and $N=8000$ for \emph{letters}), including the initial set. After the initial clean data batch, noisy data arrives in batches of $50$ instances. For the classifier with the UCI datasets we use SVM.

To synthesize the crowd labels, we evaluated \NNMC using the following parameters: number of workers $K \in \{6,8,10\}$, empty proportion $e \in \{0.1,0.2,0.3\}$, and noise rate $\varepsilon \in \{0.4, 0.6, 0.8\}$ with all four noise patterns mentioned above. Note that the empty proportion indicates the proportion of missing labels for each worker. Moreover, the mentioned noise rates $\varepsilon$ are the average of the noise rates of all the workers. The noise rates of each worker are randomly selected in the range of 10\% to 90\% with respect to the average of $\varepsilon$.

%since it is a well studied model to be used with active learning.
%The code is written in Python using the multi-class SVM in \textit{scikit-learn}~\cite{pedregosa2011scikit}. %Since SVM is not a probabilistic model, to map the decision function to probabilities for entropy calculation, function “predict_proba” uses Platt scaling  to calibrate the probabilities by using a logistic regression on SVM scores.
Our algorithm queries the true label of the $r=5$ most informative noisy samples per batch via the oracle. We repeat each experiment 50 times and report the average final accuracy computed on the test set.

\subsubsection{Networks Architecture and Training}
For \NNMC, we use a Multi-layer Perceptron (MLP) with two hidden layers with $64$ and $32$ neurons respectively, and \textit{tanh} activation function, where the input layer size corresponds to the number of workers. For \MasMet, for CIFAR-10 we consider a CNN architecture which consists of 6 convolutional layers followed by 2 fully connected layers% for multiple annotators experiments
~\cite{wang2018iterative}. The activation function is ReLU. To estimate each annotator confusion matrix, our DNN is trained for 130 epochs using SGD optimizer with momentum 0.9, weight decay $10^{-4}$, learning rate 0.01, and mini-batch size of 128 instances.

\subsection{Confusion Matrix Comparison}
Here we compare the performance of off-line \NNMC and \MasMet. In Table~\ref{tab:cm_res} reports the error rate of aggregated labels for \NNMC and \MasMet with different number of annotators (workers). We vary the noise rate $\varepsilon \in \{0.4, 0.6, 0.8\}$, which this value is the mean of noise over the annotators for the \textit{uniform} noise pattern. Furthermore, we test the effect of mixed noise pattern, where the annotators noise patterns are different from each other while the average noise rate is $\varepsilon=0.6$ (termed mixed pattern). We set the patterns with $10$ workers based on the following sequence: [\textit{bimodal}, \textit{truncnorm}, \textit{flip}, \textit{uniform}, \textit{bimodal}, \textit{truncnorm}, \textit{flip}, \textit{uniform}, \textit{bimodal}, \textit{truncnorm}].  For the case of 6 and 8 workers, we use the first 6 and 8 patterns respectively. Across the experiment in Table~\ref{tab:cm_res}, \MasMet obtains a better error rate than \NNMC. It shows that a small proportion of trusted data with clean samples can improve the estimation of the confusion matrix. The difference between these two models becomes larger by increasing the noise rate. In other words, when the number of annotators is 6, for $\varepsilon = 0.4$ and $\varepsilon = 0.8$, the difference is 1.37 and 12.07, respectively. In addition, increasing the number of annotators reduces the error rate as it increase the chance of extracting the true label. Moreover, an interesting observation is on the effect of the mixed noise pattern. As the table shows, having a mixed pattern results in lower error rates compared to the case of all \textit{uniform} noises.

\begin{table}[t]
\caption{Error rates (\%) for different confusion matrix estimation methods with empty proportion of 0.1, and different noise rates and patterns.}
\label{tab:cm_res}
\resizebox{\linewidth}{!}{%
\begin{tabular}{c|c|c|c|c}
\hline 
\rowcolor{gray!25}
\multicolumn{5}{c}{\textit{\# of Workers = 6}} \\ \hline \hline 
\rowcolor{gray!25}
\textbf{Method} & \textbf{Uniform 0.4} & \textbf{Uniform 0.6} & \textbf{Uniform 0.8} & \textbf{Mixed Patterns 0.6} \\ \hline
\NNMC  & 7.47        & 29.61       & 67.77       & 18.75                 \\ \hline
\MasMet &  6.10               &  28.33               &   55.70              &  8.28                            \\ \hline \hline
\rowcolor{gray!25}\multicolumn{5}{c}{\textit{\# of Workers = 8}} \\ \hline \hline
\NNMC  &     3.77    &    21.76    &   63.72     &         12.34         \\ \hline
\MasMet &         3.66        &      21.73           &     54.34            &  4.90                            \\ \hline \hline
\rowcolor{gray!25}\multicolumn{5}{c}{\textit{\# of Workers = 10}} \\ \hline \hline
\NNMC  &    1.77     &    16.77    &   60.42     &         8.08         \\ \hline
\MasMet &         2.88        &      19.37           &     53.99            &   4.51                           \\ \hline
\end{tabular}
}
\end{table}

\subsection{\NNMC Performance on UCI Dataset}

\begin{figure*}
\begin{center}
%\subfigure[Mushrooms]{
%    \includegraphics[scale=0.4]{mushrooms-ana2d.pdf}
%}
\subfigure[USPS]{
    \includegraphics[scale=0.4]{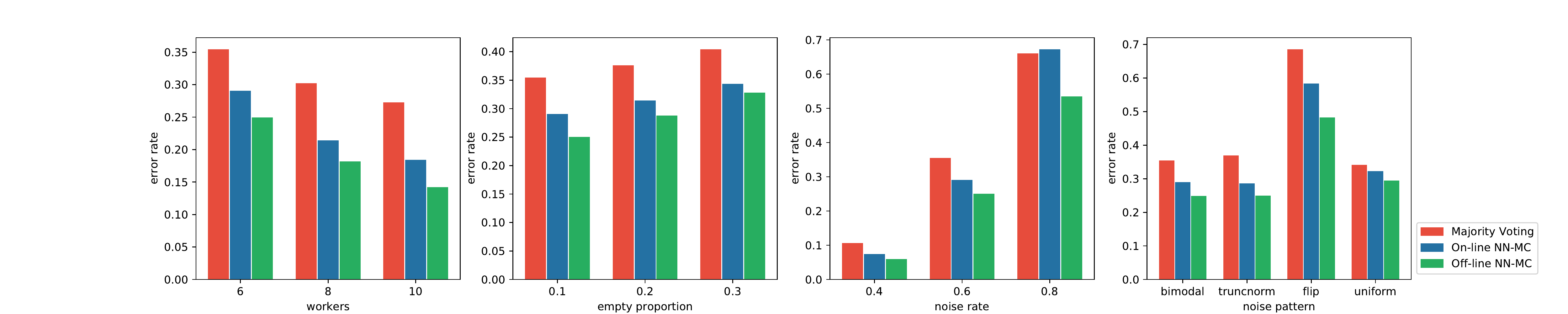}
}
\subfigure[Optdigits]{
    \includegraphics[scale=0.4]{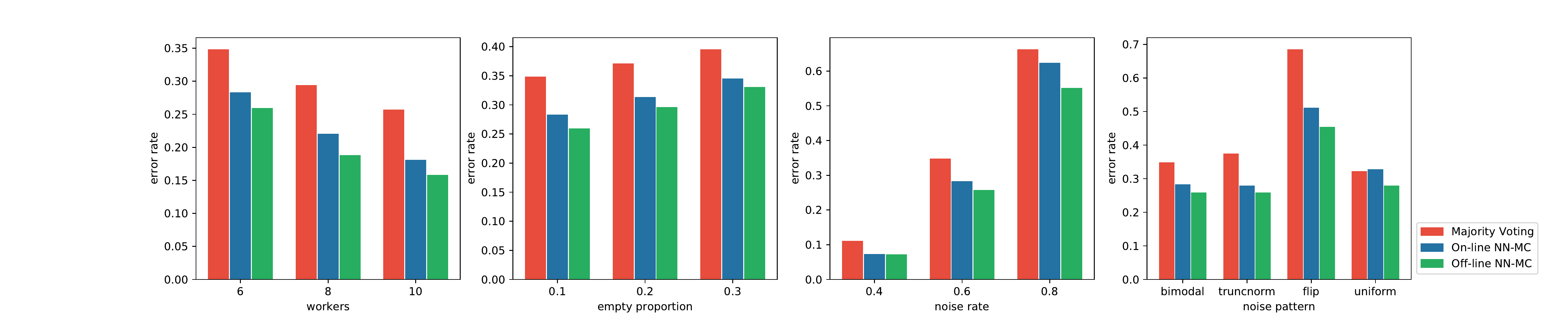}
}

%\subfigure[Letters]{
    %\includegraphics[scale=0.4]{cifar10-ana2d.pdf}
%}
\end{center}
\caption{Error rates for on-line and off-line \NNMC, and Majority Voting for \emph{usps} and \emph{optdigits}. At each row, one parameter is changing while the other parameter are fixed. The values for fixed parameters are: number of workers $K=6$, empty proportion $e=0.1$, noise rate $\varepsilon=0.6$, noise pattern bimodal.}
\label{ana2d1}
\end{figure*}

\begin{figure*}
\begin{center}
%\subfigure[Mushrooms]{
%    \includegraphics[scale=0.4]{mushrooms-ana2d.pdf}
%}

\subfigure[Pendigits]{
    \includegraphics[scale=0.4]{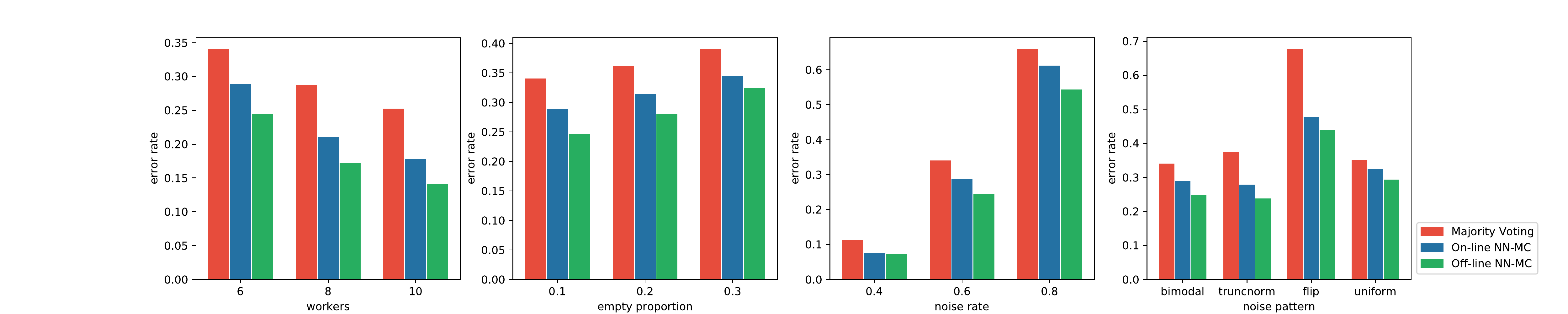}
}
\subfigure[Letters]{
    \includegraphics[scale=0.4]{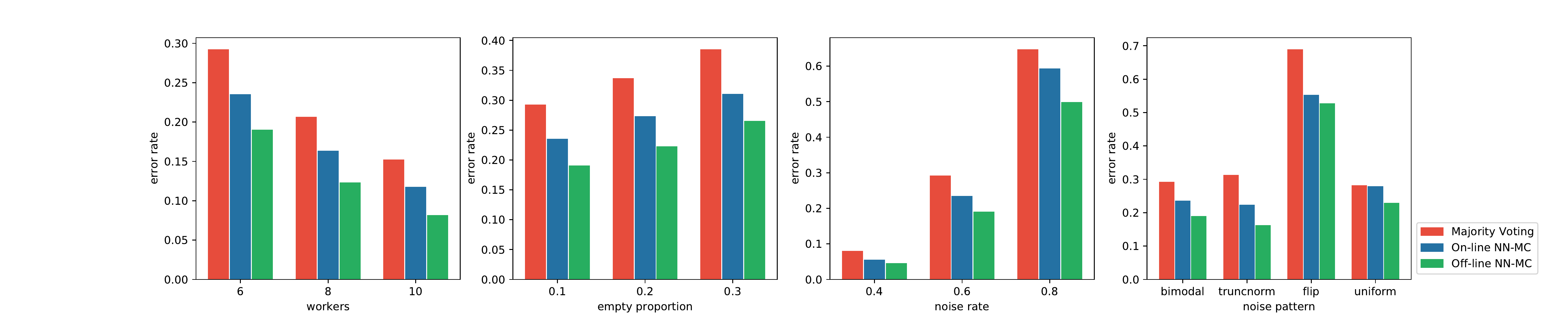}
}

%\subfigure[Letters]{
    %\includegraphics[scale=0.4]{cifar10-ana2d.pdf}
%}
\end{center}
\caption{Error rates for on-line and off-line \NNMC, and Majority Voting for \emph{pendigits} and \emph{optdigits}. At each row, one parameter is changing while the other parameter are fixed. The values for fixed parameters are: number of workers $K=6$, empty proportion $e=0.1$, noise rate $\varepsilon=0.6$, noise pattern bimodal.}
\label{ana2d2}
\end{figure*}

We extensively analyze the performance of on-line \NNMC and off-line \NNMC under different settings of number of workers, empty proportions, noise rates and noise patterns in Figure~\ref{ana2d1} and~\ref{ana2d2}. We investigate the influence of every single parameter by changing one parameter value while fixing the other parameter values. The fixed parameters are $k=6$, $e=0.1$, $\varepsilon=0.6$, and the noise pattern~\textit{bimodal}. As a baseline, we use Majority Voting and compare its performance with \NNMC. 

\begin{table}[t]
\caption{Label Error-rates (\%) for On-Line Label Aggregation, and Final Accuracy with Active Learning for $r=5$, $K=6$, $e=0.1$, $\varepsilon=0.6$ and \textit{bimodal} noise pattern.}
\label{error-rates}
\resizebox{\linewidth}{!}{%
\begin{tabular}
{c|cccc}\hline
\rowcolor{gray!25} \textbf{Method} & \textbf{USPS} & \textbf{Optdigits} & \textbf{Pendigits} & \textbf{Letters} \\ \hline
%Majority Voting & 35.47 & 34.85 & 34.04 & 29.26 \\ \hline
%Off-line \NNMC & 24.89 & 25.76 & 24.53 & 19.05  \\ \hline
Error Rate On-line \NNMC & 29.04 & 28.34 & 28.85 & 23.51  \\ \Xhline{1pt} 
Accuracy On-line \NNMC+AL (\emph{LC}) & 89.64 & 88.89 & 89.94  & 86.20  \\ \hline
Accuracy On-line \NNMC+AL (\emph{BvSB}) & 96.12 & 92.36 & 91.62  & 88.24  \\ \hline
\end{tabular}
}

\end{table}

% In Fig~\ref{ana2d1} and~\ref{ana2d2},
We can see that label aggregation methods can achieve higher accuracy when we have more workers to provide potentially noisy labels for each instance (see plots in first column). The reason is that more eligible workers for each instance corresponds to more information to correctly calculate the data likelihood and make a more precise prediction. %For Mushrooms, we can see that increasing the number of workers does not work well, because the workers are somewhat unqualified. Mushrooms only has 2 classes, but the workers even have a 0.6 noisy rate. It's obvious too high. 
According to the plots in the second column, we can see that a lower empty proportion will increase the accuracy of all methods, because the lower empty proportion represents more labels for each instance on average. The plots in the third column show that lower noise rates of the potentially noisy labels can help the label aggregation algorithms make a better prediction. In the last column, the \textit{flip} noisy pattern has the worst accuracy for all datasets. Comparing to other patterns, for each true label class (each row in the confusion matrix), \textit{flip}'s probability mass concentrate on one single wrong label. This phenomenon will significantly disturb the label aggregation algorithm to correctly learn the confusion matrices of the workers according to the potentially noisy labels. %In the last figure of Mushrooms, for the \textit{flip} and \textit{uniform} pattern, we see that \NNMC methods perform worse than Majority Voting. Actually, our on-line \NNMC and off-line \NNMC are both initialize by majority voting. Therefore, if the error rate of Majority Voting is extremely high (e.g. 70\%), the \NNMC algorithm may stop at a bad local optima. However, this is a rare situation.

\subsection{Results on Relabeling with Active Learning}
We study the effect of active learning to identify and relabel wrong aggregated labels after applying on-line \NNMC. For each batch of data that arrives, first we find the aggregated labels and then further cleanse them by applying active learning with \emph{LC} and \emph{BvSB} to relabel the informative data. Table~\ref{error-rates} shows the error rate for on-line \NNMC, and the effect on accuracy of incorporating active relabeling after \NNMC with the fixed parameters used in the previous section. As the results show, active relabeling helps in achieving a high accuracy by only relabeling $10\%$ of data instances per batch. Among the datasets, \emph{letters} seems to be more difficult to classify since it has more classes, although label aggregation succeeds in estimating a more clean label set. Among the active learning methods, \emph{BvSB} performs better in selecting the informative data, since it focuses on the two top classes, whereas \emph{LC} considers only the highest probable class.

\section{Conclusion}
In this paper we address the challenges and solutions of how to design an end-to-end learning framework from noisy crowd-sourced data, with special focus on  on-line scenarios. % We review the prior art in this field and summarize their strengths and weaknesses.
We illustrate the challenges arising with on-line label aggregation of multiple workers. We propose a visionary framework which incrementally combines noisy data, expert relabelling, and supervised models for better learning results. We introduce a method to estimate the expertise of multiple annotators by estimating their confusion matrix while leveraging a small clean dataset.
To increase the quality of the labels and benefit from an expert labeler, we relabel suspiciously noisy aggregated labels in an efficient manner.  Our results show that the proposed label aggregation can successfully lower the labeling error rate by more than $30\%$, while relabeling only $10\%$ of the most informative samples, which results in a highly accurate classification model.% Moreover, we show that confusion matrix estimation using the trusted clean data results in up to $12\%$ labeling error rate reduction in higher noise rates.

\section{Acknowledgment}
This work has been partly funded by the Swiss National
Science Foundation NRP75 project $407540\_167266$.

\bibliographystyle{ieeetr}
\bibliography{ref}

\begin{thebibliography}{10}

\bibitem{Wang:2017:IEEETrans:ActiveImageClassification}
K.~Wang, D.~Zhang, Y.~Li, R.~Zhang, and L.~Lin, ``Cost-effective active
  learning for deep image classification,'' {\em {IEEE} Trans. Circuits Syst.
  Video Techn.}, vol.~27, no.~12, pp.~2591--2600, 2017.

\bibitem{Krawczyk:2017:activeRecognition}
B.~Krawczyk, ``Active and adaptive ensemble learning for online activity
  recognition from data streams,'' {\em Knowl. Based Syst.}, vol.~138,
  pp.~69--78, 2017.

\bibitem{Smailovic:2014:InfSci:activeSentiment}
J.~Smailovic, M.~Grcar, N.~Lavrac, and M.~Znidarsic, ``Stream-based active
  learning for sentiment analysis in the financial domain,'' {\em Inf. Sci.},
  vol.~285, pp.~181--203, 2014.

\bibitem{Zhang17:ICLR:memorization}
C.~Zhang, S.~Bengio, M.~Hardt, B.~Recht, and O.~Vinyals, ``Understanding deep
  learning requires rethinking generalization,'' in {\em {ICLR}},
  OpenReview.net, 2017.

\bibitem{Yuen:2011:passat:crowdsurvey}
M.~Yuen, I.~King, and K.~Leung, ``A survey of crowdsourcing systems,'' in {\em
  PASSAT,}, pp.~766--773, {IEEE} Computer Society, 2011.

\bibitem{Howe:2008:crowd}
J.~Howe, ``Crowdsourcing: Why the power of the crowd is driving the future of
  business,'' 2008.

\bibitem{Georgescu12:CIKM:crowdError}
M.~Georgescu, D.~D. Pham, C.~S. Firan, W.~Nejdl, and J.~Gaugaz, ``Map to humans
  and reduce error: crowdsourcing for deduplication applied to digital
  libraries,'' in {\em CIKM} (X.~Chen, G.~Lebanon, H.~Wang, and M.~J. Zaki,
  eds.), pp.~1970--1974, {ACM}, 2012.

\bibitem{Raykar10:journalOfmachine:learningCrowd}
V.~C. Raykar, S.~Yu, L.~H. Zhao, G.~H. Valadez, C.~Florin, L.~Bogoni, and
  L.~Moy, ``Learning from crowds,'' {\em J. Mach. Learn. Res.}, vol.~11,
  pp.~1297--1322, 2010.

\bibitem{Huang:2017:IJCAI:DiverseLabelers}
S.~Huang, J.~Chen, X.~Mu, and Z.~Zhou, ``Cost-effective active learning from
  diverse labelers,'' in {\em {IJCAI}} (C.~Sierra, ed.), pp.~1879--1885,
  ijcai.org, 2017.

\bibitem{Demartini12:WWW:ZenCrowd}
G.~Demartini, D.~E. Difallah, and P.~Cudr{\'{e}}{-}Mauroux, ``Zencrowd:
  leveraging probabilistic reasoning and crowdsourcing techniques for
  large-scale entity linking,'' in {\em WWW} (A.~Mille, F.~L. Gandon,
  J.~Misselis, M.~Rabinovich, and S.~Staab, eds.), pp.~469--478, {ACM}, 2012.

\bibitem{Tanno:2019:CVPR:MultiAnnotatorConfusion}
R.~Tanno, A.~Saeedi, S.~Sankaranarayanan, D.~C. Alexander, and N.~Silberman,
  ``Learning from noisy labels by regularized estimation of annotator
  confusion,'' in {\em {CVPR}}, pp.~11244--11253, 2019.

\bibitem{goldberger2017adaptation}
J.~Goldberger and E.~Ben{-}Reuven, ``Training deep neural-networks using a
  noise adaptation layer,'' in {\em ICLR}, 2017.

\bibitem{song2019selfie}
H.~Song, M.~Kim, and J.-G. Lee, ``Selfie: Refurbishing unclean samples for
  robust deep learning,'' in {\em ICML}, pp.~5907--5915, 2019.

\bibitem{yang:2018:corr:alexa}
J.~Yang, T.~Drake, A.~C. Damianou, and Y.~Maarek, ``Leveraging crowdsourcing
  data for deep active learning - an application: Learning intents in alexa,''
  {\em CoRR}, vol.~abs/1803.04223, 2018.

\bibitem{Whitehill09:NIPS:vote}
J.~Whitehill, P.~Ruvolo, T.~Wu, J.~Bergsma, and J.~R. Movellan, ``Whose vote
  should count more: Optimal integration of labels from labelers of unknown
  expertise,'' in {\em NeurIPS} (Y.~Bengio, D.~Schuurmans, J.~D. Lafferty,
  C.~K.~I. Williams, and A.~Culotta, eds.), pp.~2035--2043, Curran Associates,
  Inc., 2009.

\bibitem{settles2009active}
B.~Settles, ``Active learning literature survey,'' tech. rep., University of
  Wisconsin-Madison Department of Computer Sciences, 2009.

\bibitem{ghiassi2019robust}
A.~Ghiassi, T.~Younesian, Z.~Zhao, R.~Birke, V.~Schiavoni, and L.~Y. Chen,
  ``Robust (deep) learning framework against dirty labels and beyond,'' in {\em
  TPS-ISA}, pp.~236--244, IEEE, 2019.

\bibitem{vzliobaite2013active}
I.~{\v{Z}}liobait{\.e}, A.~Bifet, B.~Pfahringer, and G.~Holmes, ``Active
  learning with drifting streaming data,'' {\em IEEE Trans. Neural Netw. Learn.
  Syst.}, vol.~25, no.~1, pp.~27--39, 2013.

\bibitem{Younesian20:arxiv:QActor}
T.~Younesian, Z.~Zhao, A.~Ghiassi, R.~Birke, and L.~Y. Chen, ``Qactor: On-line
  active learning for noisy labeled stream data,'' {\em CoRR},
  vol.~abs/2001.10399, 2020.

\bibitem{schohn2000less}
G.~Schohn and D.~Cohn, ``Less is more: Active learning with support vector
  machines,'' in {\em ICML}, pp.~839--846, 2000.

\bibitem{holub2008entropy}
A.~Holub, P.~Perona, and M.~C. Burl, ``Entropy-based active learning for object
  recognition,'' in {\em {CVPR} Workshops}, pp.~1--8, IEEE, 2008.

\bibitem{ijcai2019haussmann}
M.~Hau{\ss}mann, F.~A. Hamprecht, and M.~Kandemir, ``Deep active learning with
  adaptive acquisition,'' in {\em {IJCAI}}, pp.~2470--2476, 2019.

\bibitem{Fu:2018:sigkdd:errorReduction}
W.~Fu, M.~Wang, S.~Hao, and X.~Wu, ``Scalable active learning by approximated
  error reduction,'' in {\em SIGKDD} (Y.~Guo and F.~Farooq, eds.),
  pp.~1396--1405, {ACM}, 2018.

\bibitem{Song:2018:Knowl.BasedSyst.:confidenceCrowd}
J.~Song, H.~Wang, Y.~Gao, and B.~An, ``Active learning with confidence-based
  answers for crowdsourcing labeling tasks,'' {\em Knowl. Based Syst.},
  vol.~159, pp.~244--258, 2018.

\bibitem{Zhong:2015:AAAI:unsure}
J.~Zhong, K.~Tang, and Z.~Zhou, ``Active learning from crowds with unsure
  option,'' in {\em Proceedings of the Twenty-Fourth International Joint
  Conference on Artificial Intelligence, {IJCAI} 2015, Buenos Aires, Argentina,
  July 25-31, 2015} (Q.~Yang and M.~J. Wooldridge, eds.), pp.~1061--1068,
  {AAAI} Press, 2015.

\bibitem{Yan11:ICML:activeCrowd}
Y.~Yan, R.~Rosales, G.~Fung, and J.~G. Dy, ``Active learning from crowds,'' in
  {\em Proceedings of the 28th International Conference on Machine Learning,
  {ICML} 2011, Bellevue, Washington, USA, June 28 - July 2, 2011} (L.~Getoor
  and T.~Scheffer, eds.), pp.~1161--1168, Omnipress, 2011.

\bibitem{Murilo2019:kdd:streaming}
H.~M. Gomes, J.~Read, A.~Bifet, J.~P. Barddal, and J.~Gama, ``Machine learning
  for streaming data: state of the art, challenges, and opportunities,'' {\em
  {SIGKDD} Explor.}, vol.~21, no.~2, pp.~6--22, 2019.

\bibitem{Lu2020:CoRR:conceptDrift}
J.~Lu, A.~Liu, F.~Dong, F.~Gu, J.~Gama, and G.~Zhang, ``Learning under concept
  drift: {A} review,'' {\em CoRR}, vol.~abs/2004.05785, 2020.

\bibitem{SahooPLH18IJCAI:fly}
D.~Sahoo, Q.~Pham, J.~Lu, and S.~C.~H. Hoi, ``Online deep learning: Learning
  deep neural networks on the fly,'' in {\em IJCAI}, pp.~2660--2666, 2018.

\bibitem{neucom18losing}
V.~Losing, B.~Hammer, and H.~Wersing, ``Incremental on-line learning: {A}
  review and comparison of state of the art algorithms,'' {\em Neurocomputing},
  vol.~275, pp.~1261--1274, 2018.

\bibitem{zhu2006effective}
X.~Zhu, X.~Wu, and Y.~Yang, ``Effective classification of noisy data streams
  with attribute-oriented dynamic classifier selection,'' {\em Knowledge and
  Information Systems}, vol.~9, no.~3, pp.~339--363, 2006.

\bibitem{chu2004adaptive}
F.~Chu, Y.~Wang, and C.~Zaniolo, ``An adaptive learning approach for noisy data
  streams,'' in {\em ICDM}, pp.~351--354, IEEE, 2004.

\bibitem{mozafari:2014:lvbd:scaling}
B.~Mozafari, P.~Sarkar, M.~J. Franklin, M.~I. Jordan, and S.~Madden, ``Scaling
  up crowd-sourcing to very large datasets: {A} case for active learning,''
  {\em Proc. {VLDB} Endow.}, vol.~8, no.~2, pp.~125--136, 2014.

\bibitem{hendrycks2018using}
D.~Hendrycks, M.~Mazeika, D.~Wilson, and K.~Gimpel, ``Using trusted data to
  train deep networks on labels corrupted by severe noise,'' in {\em NIPS},
  pp.~10456--10465, 2018.

\bibitem{Ghiassi20:arxiv:TrustNet}
A.~Ghiassi, T.~Younesian, R.~Birke, and L.~Y. Chen, ``Trustnet: Learning from
  trusted data against (a)symmetric label noise,'' {\em CoRR},
  vol.~abs/2007.06324, 2020.

\bibitem{ghiassi2020expertnet}
A.~Ghiassi, R.~Birke, R.~Han, and L.~Y. Chen, ``Expertnet: Adversarial learning
  and recovery against noisy labels,'' 2020.

\bibitem{sukhbaatar2014training}
S.~Sukhbaatar, J.~Bruna, M.~Paluri, L.~Bourdev, and R.~Fergus, ``Training
  convolutional networks with noisy labels,'' {\em ICLR Workshops}, 2015.

\bibitem{thekumparampil2018robustness}
K.~K. Thekumparampil, A.~Khetan, Z.~Lin, and S.~Oh, ``Robustness of conditional
  gans to noisy labels,'' in {\em NIPS}, pp.~10271--10282, 2018.

\bibitem{patrini2017making}
G.~Patrini, A.~Rozza, A.~Krishna~Menon, R.~Nock, and L.~Qu, ``Making deep
  neural networks robust to label noise: A loss correction approach,'' in {\em
  CVPR}, pp.~1944--1952, 2017.

\bibitem{Han:2018:NeurIPS:Masking}
B.~Han, J.~Yao, G.~Niu, M.~Zhou, I.~W. Tsang, Y.~Zhang, and M.~Sugiyama,
  ``Masking: {A} new perspective of noisy supervision,'' in {\em {NIPS}},
  pp.~5841--5851, 2018.

\bibitem{dawid1979maximum}
A.~P. Dawid and A.~M. Skene, ``Maximum likelihood estimation of observer
  error-rates using the em algorithm,'' {\em Applied statistics}, pp.~20--28,
  1979.

\bibitem{kim2012bayesian}
H.-C. Kim and Z.~Ghahramani, ``Bayesian classifier combination,'' in {\em
  Artificial Intelligence and Statistics}, pp.~619--627, 2012.

\bibitem{zhou2012learning}
D.~Zhou, S.~Basu, Y.~Mao, and J.~C. Platt, ``Learning from the wisdom of crowds
  by minimax entropy,'' in {\em Advances in Neural Information Processing
  Systems}, pp.~2195--2203, 2012.

\bibitem{zhou2014aggregating}
D.~Zhou, Q.~Liu, J.~Platt, and C.~Meek, ``Aggregating ordinal labels from
  crowds by minimax conditional entropy,'' in {\em International Conference on
  Machine Learning}, pp.~262--270, 2014.

\bibitem{zhou2016crowdsourcing}
Y.~Zhou and J.~He, ``Crowdsourcing via tensor augmentation and completion.,''
  in {\em IJCAI}, pp.~2435--2441, 2016.

\bibitem{jongen2014relationship}
P.~J. Jongen, K.~Wesnes, B.~van Geel, P.~Pop, E.~Sanders, H.~Schrijver, L.~H.
  Visser, H.~J. Gilhuis, L.~G. Sinnige, A.~M. Brands, {\em et~al.},
  ``Relationship between working hours and power of attention, memory, fatigue,
  depression and self-efficacy one year after diagnosis of clinically isolated
  syndrome and relapsing remitting multiple sclerosis,'' {\em PloS one},
  vol.~9, no.~5, p.~e96444, 2014.

\bibitem{pencavel2015productivity}
J.~Pencavel, ``The productivity of working hours,'' {\em The Economic Journal},
  vol.~125, no.~589, pp.~2052--2076, 2015.

\bibitem{Li:2013:cvpr:adaptiveActive}
X.~Li and Y.~Guo, ``Adaptive active learning for image classification,'' in
  {\em 2013 {IEEE} Conference on Computer Vision and Pattern Recognition,
  Portland, OR, USA, June 23-28, 2013}, pp.~859--866, {IEEE} Computer Society,
  2013.

\bibitem{Chaudhuri:NIPs16:ImperfectLablers}
S.~Yan, K.~Chaudhuri, and T.~Javidi, ``Active learning from imperfect
  labelers,'' in {\em NeurIPs}, pp.~2128--2136, 2016.

\bibitem{Sheng:2008:SIGKDD:Multilabeler}
V.~S. Sheng, F.~J. Provost, and P.~G. Ipeirotis, ``Get another label? improving
  data quality and data mining using multiple, noisy labelers,'' in {\em {ACM}
  {SIGKDD}}, pp.~614--622, 2008.

\bibitem{Zhang:2015:TransCybernetics:UnbalancedMultiLabeler}
J.~Zhang, X.~Wu, and V.~S. Sheng, ``Active learning with imbalanced multiple
  noisy labeling,'' {\em {IEEE} Trans. Cybernetics}, vol.~45, no.~5,
  pp.~1081--1093, 2015.

\bibitem{Zheng:2010:ICDM:ActiveCost}
Y.~Zheng, S.~D. Scott, and K.~Deng, ``Active learning from multiple noisy
  labelers with varied costs,'' in {\em {ICDM} 2010, The 10th {IEEE}
  International Conference on Data Mining, Sydney, Australia, 14-17 December
  2010} (G.~I. Webb, B.~Liu, C.~Zhang, D.~Gunopulos, and X.~Wu, eds.),
  pp.~639--648, {IEEE} Computer Society, 2010.

\bibitem{Donmez:2008:CIKM:proactiveCost}
P.~Donmez and J.~G. Carbonell, ``Proactive learning: cost-sensitive active
  learning with multiple imperfect oracles,'' in {\em CIKM} (J.~G. Shanahan,
  S.~Amer{-}Yahia, I.~Manolescu, Y.~Zhang, D.~A. Evans, A.~Kolcz, K.~Choi, and
  A.~Chowdhury, eds.), pp.~619--628, {ACM}, 2008.

\bibitem{younesian2020active}
T.~Younesian, D.~Epema, and L.~Y. Chen, ``Active learning for noisy data
  streams using weak and strong labelers,'' 2020.

\bibitem{Teh:2010:AISTATS:bit}
Y.~Yan, R.~Rosales, G.~Fung, M.~Schmidt, G.~H. Valadez, L.~Bogoni, L.~Moy, and
  J.~G. Dy, ``Modeling annotator expertise: Learning when everybody knows a bit
  of something,'' in {\em {AISTATS}} (Y.~W. Teh and D.~M. Titterington, eds.),
  vol.~9 of {\em {JMLR} Proceedings}, pp.~932--939, 2010.

\bibitem{Chaudhuri:NeurIPs15:StrongWeakLablers5}
C.~Zhang and K.~Chaudhuri, ``Active learning from weak and strong labelers,''
  in {\em NeurIPs}, pp.~703--711, 2015.

\bibitem{Zhao:PASSAT11:IncRelabel}
L.~Zhao, G.~Sukthankar, and R.~Sukthankar, ``Incremental relabeling for active
  learning with noisy crowdsourced annotations,'' in {\em PASSAT},
  pp.~728--733, {IEEE} Computer Society, 2011.

\bibitem{Yang:SIGKDD12:schoolOFthoughts}


\bibitem{Liu:2015:SIGMETRICS:activeImproving}
Y.~Liu and M.~Liu, ``An online learning approach to improving the quality of
  crowd-sourcing,'' in {\em Proceedings of the 2015 {ACM} {SIGMETRICS}}
  (B.~Lin, J.~J. Xu, S.~Sengupta, and D.~Shah, eds.), pp.~217--230, {ACM},
  2015.

\bibitem{tieleman2012lecture}
T.~Tieleman and G.~Hinton, ``Lecture 6.5-rmsprop, coursera: Neural networks for
  machine learning,'' {\em University of Toronto, Tech. Rep}, 2012.

\bibitem{joshi:2009:CVPR:multi}
A.~J. Joshi, F.~Porikli, and N.~Papanikolopoulos, ``Multi-class active learning
  for image classification,'' in {\em IEEE CVPR}, pp.~2372--2379, 2009.

\bibitem{Dua:2017}
D.~Dua and C.~Graff, ``{UCI} machine learning repository,'' 2017.

\bibitem{kriz-cifar10}
A.~Krizhevsky, V.~Nair, and G.~Hinton, ``Cifar-10 (canadian institute for
  advanced research),''

\bibitem{goodfellow:2015:explaining}
I.~J. Goodfellow, J.~Shlens, and C.~Szegedy, ``Explaining and harnessing
  adversarial examples,'' in {\em {ICLR}}, 2015.

\bibitem{wang2019symmetric}
Y.~Wang, X.~Ma, Z.~Chen, Y.~Luo, J.~Yi, and J.~Bailey, ``Symmetric cross
  entropy for robust learning with noisy labels,'' in {\em IEEE ICCV},
  pp.~322--330, 2019.

\bibitem{wang2018iterative}
Y.~Wang, W.~Liu, X.~Ma, J.~Bailey, H.~Zha, L.~Song, and S.-T. Xia, ``Iterative
  learning with open-set noisy labels,'' in {\em IEEE {CVPR}}, pp.~8688--8696,
  2018.

\end{thebibliography}

\end{document}